# Prognostic & Health Management (PHM) Tool for Robot Operating System (ROS)

A PREPRINT

Hakan Gencturk[1]
hakan.gencturk@inovasyonmuhendislik.com

Elcin Erdogan[1]
elcin.erdogan@inovasyonmuhendislik.com

Mustafa Karaca[1]
mustafa.karaca@inovasyonmuhendislik.com

Dr. Ugur Yayan[1]
ugur.yayan@inovasyonmuhendislik.com

November 18, 2020

## ABSTRACT

Nowadays, prognostics-aware systems are increasingly used in many systems and it is critical for sustaining autonomy. All engineering systems, especially robots, are not perfect. Absence of failures in a certain time is the perfect system and it is impossible practically. In all engineering works, we must try to predict or minimize/prevent failures in the system. Failures in the systems are generally unknown, so prediction of these failures and reliability of the system is made by prediction process. Reliability analysis is important for the improving the system performance, extending system lifetime, etc. Prognostic and Health Management (PHM) includes reliability, safety, predictive fault detection / isolation, advanced diagnostics / prognostics, component lifecycle tracking, health reporting and information management, etc. This study proposes an open source robot prognostic and health management tool using model-based methodology namely" Prognostics and Health Management tool for ROS". This tool is a generic tool for using with any kind of robot (mobile robot, robot arm, drone etc.) with compatible with ROS. Some features of this tool are managing / monitoring robots' health, RUL, probability of task completion (PoTC) etc. User is able to enter the necessary equations and components information (hazard rates, robot configuration etc.) to the PHM tool and the other sensory data like temperature, humidity, pressure, load etc. In addition to these, a case study is conducted for the mobile robots (OTA) using this tool.

## 1. Introduction

It is indicated that human employees add significant cost to the company with training and memorizing the tasks. However physical robot does not add additional or incremental cost in operations.[3] With decreasing cost in data computing and electronics and AI-based algorithms in executing tasks, robots have started having advantage against human employees both in cognitive and analytical works. On the other side, beyond the development, sustainability of these robots are critical in operations. Robots which have critical role in operations can lead significant expenses to the companies. Unknown errors and unexpected failures can restrain operations or cause work losses in production. In this manner, Prognostics and Health Management (PHM) have become important tool for maintaining effective operations.

Reliability generally shows how successful a system is in performing its intended function. In order to increase the system performance, intensive studies are carried out on reliability models and analytical tools to reveal system health. It is true that the consequences of unreliability in engineering can be very costly and often tragic. In a changing world, reliability becomes more important to ensure that systems operate with high performance and gain functionality. Reliability engineering is applied in many fields such as defense industry, robotics industry, aerospace industry etc. The current generation of mobile robots have poor reliability. In order to design more reliable robots, we need analytical tools for predicting robot failure. The shift towards the Robot centered production also creates demand

[1] Research and Development Department, Inovasyon Muhendislik Ltd. Sti., Eskisehir, Turkey

for highly accurate, reliable and cost-efficient robot solutions and importance of the robots increase in industrial applications. Reliability and maintenance of these robots are getting more critical. Especially, with emerging complex applications in factory floor like robot-robot operations and human-robot operations, it is important to know reliability and status of robot. While Human-Robot interactions are increasing with building complex systems, It is indicated that more complex operations create more reasons faults and failures to occur [1,2]. These faults and failures can cause unexpected loss in efficiency, quality and productivity. In order to minimize these fault and failures, health of the system and all sub-systems should be monitored.

The importance of reliability of the robots and enabling PHM have stated in the US Department of Defense 5000.2 policy document [4] as a defense acquisition. This policy states that "program managers shall optimize operational readiness through affordable, integrated, embedded diagnostics and prognostics, embedded training and testing, serialized item management, automatic identification technology, and iterative technology refreshment". Also, in [41], it is indicated that maintenance is a critical part of production. Thus, Prognostic system development and enabling PHM has become a necessity in all operations. In these situations, Prognostic and health management systems have gained importance as a result. There is a great motivation in PHM plans to apply industrial robotic applications in academy and industry [42].

This study proposes an open source robot prognostic and health management tool using model-based methodology namely" Prognostics and Health Management tool for ROS". This tool is a generic tool for using with any kind of robot (mobile, arm, drone etc.) compatible with ROS. Some features of this tool are managing / monitoring robots' health, RUL, probability of task completion (PoTC) etc. User is able to enter the necessary equations and components information (hazard rates, robot configuration etc.) to the PHM tool and the other sensory data like temperature, humidity, pressure, load etc. In addition to these, a case study is conducted for the mobile robots (OTA) using this tool. In these following sections, Part 2 explains previous work on PHM. Next, Part 3 provides background and hazard rate calculation formulas for PHM tool infrastructure. Part 4 explains PHM tool and provides a guide. Part 5 introduce case study for PHM tool and Part 6 provides overall conclusion.

## 2. Previous Studies

The goal of maintenance includes 2 types of maintenance types. In literature, PHM goes beyond these applications with correct future predictions. In literature, there are methods and applications are made for estimating robot reliability and assessing RUL. In this area, early comprehensive study on PHM is made by Ahmad and Kamaruddin presents overview on time-based maintenance (TBM) and condition-based maintenance (CBM). In time-based maintenance, equipment is assumed to be have predictable failure behavior and characteristic. Time-based maintenance methods fall behind the modern technological trends in today's technology [5]. In complex technologies where components have no characteristics of failure, condition-based maintenance which is also known as predictive maintenance become most popular maintenance technique. In condition-based maintenance, trend is going towards the advanced technologies via PHM. PHM is mostly referred as preventative maintenance which is developed for component/system safety [6]. In 2006, detailed overview about PHM and its benefits is given by Vichare and Pecht [7]. In current state of PHM applications, Product PHM is more popular than Process PHM. In current status, PHM techniques are mostly developed for components or equipment. There is lacking innovation on overall system level PHM and robot reliability methods [6].

In 2002, Shetty et al. [8] presented work on remaining life assessment of a shuttle remote manipulator system. this study applied PHM methodology for end effector electronics unit of the robot arm system. Thermal and vibration loads are used for damage models and prognostics estimated that electronic units can operate another 20 years. Another prognostic study applied for aerospace domain is done by Mathew et al. [9]. In this study, prognostic remaining-life assessment task for rocket booster circuit cards is done. In study, estimation of the number of future missions that circuit card can operate which is called as virtual PHM process is depend on 5 stages. Design capture and model analysis, life cycle load history characterization, load transformation on the card, damage assessment on each card component and estimation remaining useful life of circuit card. In study use case, virtual remaining life assessment process provided outcome of the circuit can operate 39 more missions on aerospace. [9] Another study made by Nasser and Curtin [10] takes power supply failures into consideration. This study focuses PHM application on solder degradation, solder fracture and fatigue. Study introduces this PHM technique in 5 steps; 1) acquiring sensor data (temperature, vibration, Gvalue), this translates into Thermal and Vibration map. 2) Finite Element stress analysis 3) conducting status prediction for each point 4) prognostic health prediction for Power supply system. Proposed method predicts remaining life of power supply system with over 85% confidence. [10] In another study, micro-programmable module is customized for health and usage monitoring. In case study, presented method collects analyses CPU temperature, HDD temperature, ambient temperature, CPU usage and, fan condition of the laptop in use. With the acquired data, guideline for laptop damage accumulation and remaining-life estimation is proposed [11]. For the machining operation, Biehl et al.[12] proposed thin film sensor system which resistance changes are depend on various preload forces and causes systematic manufacturing errors. These measured results help monitoring of the ball screw drivers' health.[12]. In [13], a study made for detecting degradation of machine components, a method including Internal Measurement Unit (IMU) for identifying translational and angular errors in robot arm movements. For this method, accelerometer and rate gyroscope data is used. With collected data, IMU have detected micrometer and microradian level degradation on robot arm end. Writers indicates that with coupling standards and validation of the system, Industrial IMU can provide real-time equipment health prediction and prognostics.[13] In paper [14], Zhang and Han worked on reliability of robotic manipulator. Their study proposed reliability analysis method with taking randomness in link lengths and joint angles into account to prognose kinematics reliability. The proposed reliability method is based on Dimension Reduction and Saddlepoint Approximation. In industrial robotic manipulator, DR achieves good accuracy in randomness and SPA ensures final reliability. However, writers have concern in encountering real applications and these methods need future improvement [14]. Similar problem is also have taken into concern in [15]. The uncertainty in link dimension variables and joint clearances is also took account as a problem. Denavit–Hartenberg method is used in kinematic models which system assumed to be have uncertain variables. Then, Sobol sensitivity analysis method implementation showed that parameters have huge impact on positioning accuracy of end-effector and sensitivity of variable errors are analyzed. This analyses method uncovers reliability framework of robot. In experiment in simulation shows that proposed method provides reliability in position error and joint rotation errors in robot designs. In the other hand, for positioning reliability and multipoint trajectory analysis, this method is still lacking [15]. Future study made by Wu et al. [16] provides new method for comprehensive evaluation of the positional accuracy. In their novel method, external 3D position measurement of point and trajectory is obtained with sparse gid numerical integration and saddle point approximation methods. Kinematic error models are also obtained with these methods. To assess reliability of the robot kinematic accuracy, extreme value distribution of response is applied. Dissolve eigen-decomposition of covariance matrix for three coordinates of positional point is implemented for correlation three coordinates. Also, probability density function and probability of failure calculations are used for saddle point approximation. To prove effectiveness of the method, 4 different examples are proposed. However, since the method is depending on integration and weights in grid integration method. Proposed method is not applicable for various cases.[16]

There are some studies that provides system-level PHM methods. Adaptive Multi-scale PHM (AM-PHM) is one of these methods proposed for robotic assembly processes in smart manufacturing system. This method applies AM-PHM in every decision point of holistic system. Method takes job requirements in operations and reports performance and health estimates to upper level structure. Decisions are based on current and projected health states of the system. AM-PHM information domain consist of machine level, work cell level and assembly line level PHM structure. layers from low to higher levels provide machine level health to assembly Line level health information [17]. A study by Qiao et al. [18] provides quick robot health assessment methodology for enabling PHM. In this study Positional (position and orientation accuracy) health changes are taking into consideration. For the assessment of the robot health, fixed loop motion test method is developed. In this method, 7-D information (X, Y, Z, roll, pitch, yaw, and time) is analyzed. With provided error modelling, users can assess the robot positional health faster with higher accuracy.[18] Another study proposes top level and component level PHM for detecting robot performance degradation including robot tool center accuracy degradation. In this study, four dimensional PHM is presented. In study cost function works on solving robot error model to identify root causes of faults and failures. Similar to previously presented study, 7-D information is used for health analysis. [19,20] At next, in Prognostics, Health Management and Control (PHMC) project, representative manufacturing robot work cells are proposed as use case options in physical test beds for PHM systems development.[21].

There are solutions in market aiming for PHM. First one of these solutions is FlexPro. FlexPro is data analyzation and presentation program that includes fast data search, data transfer in various standards, analysis procedures for measured data and processing measurement values in multi treaded workflows. FlexPro can be used as a sub-system for developing PHM solutions. However, program itself is not capable of implementing PHM technologies. [22] Reliability workbench is a software in development since 1980s and provides block diagram reliability analysis, fault tree and common important analysis methods, standards for supporting system analysis. Reliability workbench provides accurate system reliability predictions but for the prognostic side this software is still lacking. [23] ITEM ToolKit is a reliability software includes comprehensive reliability prediction and reliability analysis modules. Software can assess maintainability and safety of electrical/mechanical components of the system. ITEM ToolKit platform offers scalable analysis with including modules can be purchased. [24] Fusce MADe is a software for enabling better decisions about designing and supporting safety in workspace and mission critical equipment. MADe provides modelling & analysis in multi-domain systems. Provides design optimization, reliability and maintainability analysis, safety and risk assessment and PHM for designing/validating diagnostic requirements for CBM. PHM method has generation of sensor sets utilizing a genetic algorithm analysis on maximum coverage. PHM module provides model-based approach, automated sensor set generation and diagnostic decision support. Also MADe provides sensor allocation and optimization. When all sensors are allocated, the parameters of each sensor are entered to enable optimization based on user inputs. MADe also provides diagnostic with identifying specific failures in system. [25] ALD's Reliability software is a suite for reliability prediction, Availability, maintainability analysis, Safety Assessment, Quality Management, Safety Management and Industrial Process Control. ALD Reliability software has MTBF Calculator, Fault Tree Analyzer and PHM commander in it. PHM commander provides failure diagnostics performs before and after failure occurs and prognostics provide early failure detection. ALD PHM solution also supports data driven approach for prediction RUL. [26] Calce PHM maintenance planning tool is a stochastic decision model which determines schedule of the maintenance times. System models enable optimal health monitoring, life consumption monitoring. This tool can be used in optimizing safety margins and determine best maintenance strategies in multiple systems. In 2007, this tool is extended to address false positives and it is performing ROI studies on single system. Tool includes computed metrics which are; life cycle cost, avoided failures and operational availability. [27]

# 3. Definitions and Preliminaries

In this section, the calculation of the hazard rate and reliability values of a robotic system and the configurations (serial, parallel, etc.) of the mechanical or electrical equipments in the robot are explained in detail. At the same time, the probability of completing a task that comes to the robot is calculated based on the reliability value of the robot. Hazard rate calculation formulas for the mechanical and electrical equipments in the robot are explained in Section 3.1. The serial/parallel configuration of the equipments in the robotic system, the reliability calculation of the robot and the Life Distribution Models used in calculating the reliability are explained in Section 3.2. And finally, the calculation of probability of the robot completion a task is explained in Section 3.3.

## 3.1 Hazard Rate Calculation of System

A robotic system consists of many equipments and with their configurations. These components can simply be the battery, power control unit, wheels, motor control unit, sensors etc. By calculating the hazard rate values of each components, the hazard rate of overall system can be calculated. Component specific hazard rate calculation formulas are used to calculate the hazard rate value of each electrical and mechanical components.

PHM Tool contains hazard rate calculation formulas for some electrical and mechanical equipments by default. These components are given in Table 1.

Table 1: Electrical and Mechanical Equipments in PHM Tool

| Electrical Equipments | Mechanical Equipments |
|---|---|
| Capacitor | Spring |
| Diode | Gears |
| Inductor | Bearing |
| Transistor | Actuators |
| Fuse | Shafts |
| Resistor | Electric Motors |
| Rotating Devices, Motors | Mechanical Couplings |
| Relays | Battery |
| Connectors, General | |
| Connectors Sockets | |
| Quartz Crystals | |

### 3.1.1 Mechanical Equipment Hazard Rate Formulation

In this section, hazard rate value calculation of used mechanical equipments in OTA is given though the all equipments in Table 1 are in PHM Tool. Most of the hazard rate calculations are taken from [28]. The study [28] contains more detailed information about the calculations.

i. **Bearings:** In OTA, bearings are used in the mobility module. The bearing failure rate can be expressed as eq. (1). Also, the variables of bearing hazard rate calculation formula are shown in Table 2:

$$\lambda_{BE} = \lambda_{BE,B} * C_Y * C_R * C_V * C_{CW} * C_t * C_{SF} * C_C \quad (1)$$

Table 2: Variables are used in calculation of bearings hazard rate

| | |
|---|---|
| $\lambda_{BE,B}$ | Base failure rate, failures/million hours |
| $C_Y$ | Multiplying factor applied load |
| $C_R$ | Life adjustment factor for reliability |
| $C_V$ | Multiplying factor for lubricant |
| $C_{CW}$ | Multiplying factor for water contaminant level |
| $C_t$ | Multiplying factor for operating temperature |
| $C_{SF}$ | Multiplying factor for operating service conditions |
| $C_C$ | Multiplying factor for lubrication contamination level |

ii. **Electric Motors:** In OTA, electric motors are used in the mobility module. The total motor system failure rate is the sum of the failure rates of each of the parts in the motor as shown in eq. (2). Also, the variables of eq. (2) are shown in Table 3:

$$\lambda_M = (\lambda_{M,B} * C_{SF}) + \lambda_{WI} + \lambda_{BS} + \lambda_{ST} + \lambda_{AS} + \lambda_{BE} + \lambda_{GR} + \lambda_C \,) \quad (2)$$

Table 3: Variables are used in calculation of electric motors hazard rate

| | |
|---|---|
| $\lambda_{M,B}$ | Base failure rate of motor, failures/million hours |
| $C_{SF}$ | Motor load service factor |
| $\lambda_{WI}$ | Failure rate of electric motor windings, failures/million hours |
| $\lambda_{BS}$ | Failure rate of brushes, 3.2 failures/million hours/brush [17] |
| $\lambda_{ST}$ | Failure rate of the stator housing, 0.001 failures/million hours [17] |
| $\lambda_{AS}$ | Failure rate of the armature shaft, failures/million hours |
| $\lambda_{BE}$ | Failure rate of bearings, failures/million hours |
| $\lambda_{GR}$ | Failure rate of gears, failures/million hours |
| $\lambda_C$ | Failure rate of capacitor (if applicable) [30] |

iii. **Battery:** The battery failure rate can be expressed as:

$$\lambda = \lambda_0 * 10^{-9}/h \quad (3)$$

The value $\lambda_0$ will be obtained using the Table 4.

Table 4: $\lambda_0$ to calculate battery hazard rate

| **Device Type** | **Type** | $\boldsymbol{\lambda_0}$ |
|---|---|---|
| Batteries: primary cells | | 20 |
| Batteries: secondary cells | Ni-Cd | 100 |
| | Li-Ion | 150 |

### 3.1.2 Electrical Equipment Hazard Rate

In this section, calculation of electrical equipment's hazard rate is given. All hazard rate calculation is taken from [30]. The study [30] contains more detailed information about the calculations. In this section, all part quality and environment factor values are depending on the Table 5 that is given below.

The quality of an electrical equipment has a direct effect on its failure rate and is seen as the $\pi_Q$ quality factor in the failure rate calculation of the equipment.

Some electrical components with their quality designators are shown in Table 5.

Table 5: Parts with Multi-Level Quality Specifications

| Part | Quality Designators |
|---|---|
| Microcircuits | SI B, B-1, Other: Quality Judged by Screening Level |
| Discrete semiconductors | JANTXV, JANTX, JAN |
| Capacitors, Established Reliability (ER) | D, C, S, R, B, P, M, L |
| Resistors, Established Reliability (ER) | S, R, P, M |
| Coils, Molded, R.F., Reliability | S, R, P, M |
| Relays, Established Reliability (ER) | R, P, M, L |

In calculating the hazard rate of electrical equipments, the environmental impacts on reliability is provided by the environmental factor, $\pi_E$. There are three ground environments in literature; benign, fixed and mobile [30].

According to Military Handbook [30] these environments explained as; "Benign grounds ($G_B$) are non-mobile, temperature and humidity-controlled environments readily accessible to maintenance; includes laboratory instruments and test equipment, medical electronic equipment, business and scientific computer complexes, and missiles and support equipment in ground silos.", "Fixed grounds ($G_F$) are moderately controlled environments such as installation in permanent racks with adequate cooling air and possible installation in unheated buildings; includes permanent installation air traffic control radar and communications facilities." and "Mobile grounds ($G_M$) are equipment installed on wheeled or tracked vehicles and equipment manually transported; includes tactical missile ground support equipment, mobile communication equipment, tactical fire direction systems, handheld communications equipment, laser designations and range finders."

i. **Capacitor:** In OTA, capacitors are used in the many modules such as power, sensing and communication. The capacitor failure rate can be expressed as eq. (4) and variables are used in capacitor hazard rate calculation formula is shown in Table 6:

$$\lambda_p = \lambda_b * \pi_T * \pi_C * \pi_V * \pi_{SR} * \pi_Q * \pi_E \; Failures/10^6 Hours \tag{4}$$

Table 6: Variables are used in calculation of capacitor hazard rate

| | |
|---|---|
| $\lambda_b$ | Base Failure Rate |
| $\pi_T$ | Temperature Factor |
| $\pi_C$ | Capacitance Factor |
| $\pi_V$ | Voltage Stress Factor |
| $\pi_{SR}$ | Series Resistance Factor |
| $\pi_Q$ | Quality Factor |
| $\pi_E$ | Environment Factor |

ii. **Diode:** In OTA, diodes are used in power module. The diode failure rate can be expressed as eq. (5) and variables of eq. (5) are shown in Table 7:

$$\lambda_p = \lambda_b * \pi_T * \pi_S * \pi_C * \pi_Q * \pi_E \; Failures/10^6 Hours \tag{5}$$

Table 7: Variables are used in calculation of diode hazard rate

| | |
|---|---|
| $\lambda_b$ | Base Failure Rate |
| $\pi_T$ | Temperature Factor |
| $\pi_S$ | Electrical Stress Factor |
| $\pi_C$ | Contact Consttuction Factor |
| $\pi_Q$ | Quality Factor |
| $\pi_E$ | Environment Factor |
| $\lambda_b$ | Base Failure Rate |

**iii. Inductor:** In OTA, inductors are used in the power module. The inductor failure rate can be expressed as eq. (6) and variables of eq. (6) are shown in Table 8:

$$\lambda_p = \lambda_b * \pi_T * \pi_Q * \pi_E \; Failures/10^6 Hours \quad (6)$$

Table 8: Variables are used in calculation of inductor hazard rate

| | |
|---|---|
| $\lambda_b$ | Base Failure Rate |
| $\pi_T$ | Temperature Factor |
| $\pi_Q$ | Quality Factor |
| $\pi_E$ | Environment Factor |

**iv. Fuse:** In OTA, fuses are used in the power and communication modules. The fuse failure rate can be expressed as eq. (7) and variables of eq. (7) are shown in Table 9:

$$\lambda_p = \lambda_b * \pi_E \; Failures/10^6 Hours \quad (7)$$

Table 9: Variables are used in calculation of fuse hazard rate

| | |
|---|---|
| $\lambda_b$ | Base Failure Rate |
| $\pi_E$ | Environment Factor |

**v. Resistor:** In OTA, resistors are used in the many modules such as power, sensing and communication. The resistor failure rate can be expressed as eq. (8) and variables of eq. (8) are shown in Table 10:

$$\lambda_p = \lambda_b * \pi_T * \pi_P * \pi_S * \pi_Q * \pi_E \; Failures/10^6 Hours \quad (8)$$

Table 10: Variables are used in calculation of resistor hazard rate

| | |
|---|---|
| $\lambda_b$ | Base Failure Rate |
| $\pi_T$ | Temperature Factor |
| $\pi_P$ | Power Factor |
| $\pi_S$ | Power Stress Factor |
| $\pi_Q$ | Quality Factor |
| $\pi_E$ | Environment Factor |

**vi. Rotating Devices, Motors:** The rotating device failure rate can be expressed as eq. (9) and variables of eq. (9) are shown in Table 11. For more details about the variables, see the reference [30]:

$$\lambda_P = \left[\frac{\lambda_1}{A\alpha_B} + \frac{\lambda_2}{B\alpha_W}\right] \times 10^6 \; Failures/10^6 \; Hours \quad (9)$$

Table 11: Variables are used in calculation of rotating devices, motors hazard rate

| | |
|---|---|
| $\alpha_B$ | Weibull Charactistic Life for the Motor Bearing |
| $\alpha_W$ | Weibull Characteristic Life for the Motor Windings |

**vii. Connectors, General:** In OTA, general connectors are used in the many modules such as power, sensing and communication. The connector (general) failure rate can be expressed as eq. (10) and variables of eq. (10) are shown in Table 12:

$$\lambda_p = \lambda_b * \pi_T * \pi_K * \pi_Q * \pi_E \ Failures/10^6 \ Hours \tag{10}$$

Table 12: Variables are used in calculation of general connectors hazard rate

| | |
|---|---|
| $\lambda_b$ | Base Failure Rate |
| $\pi_T$ | Temperature Factor |
| $\pi_K$ | Mating/Unmating Factor |
| $\pi_Q$ | Quality Factor |
| $\pi_E$ | Environment Factor |

**viii.** **Connectors Sockets:** In OTA, sockets are used in the power module. The connector (socket) rate can be expressed as eq. (11) and variables of eq. (11) are shown in Table 13:

$$\lambda_p = \lambda_b * \pi_P * \pi_Q * \pi_E \ Failures/10^6 \ Hours \tag{11}$$

Table 13: Variables are used in calculation of socket hazard rate

| | |
|---|---|
| $\lambda_b$ | Base Failure Rate |
| $\pi_P$ | Active Pins Factor |
| $\pi_Q$ | Quality Factor |
| $\pi_E$ | Environment Factor |

**ix.** **Quartz Crystals:** In OTA, quartz crystals are used in the power and communication modules. The crystal quartz failure rate can be expressed as eq. (12) and variables of eq. (12) are shown in Table 14:

$$\lambda_p = \lambda_b * \pi_Q * \pi_E \ Failures/10^6 \ Hours \tag{12}$$

Table 14: Variables are used in calculation of quartz crystals hazard rate

| | |
|---|---|
| $\lambda_b$ | Base Failure Rate |
| $\pi_Q$ | Quality Factor |
| $\pi_E$ | Environment Factor |

## 3.2 Configuration and Reliability estimation of System

In this section, calculation of the reliability of a robotic system, configuration of the system and life distribution models used when calculating reliability are explained in detail.

### 3.2.1 Configuration of System

Over the years, reliability researchers working and developed various reliability assessment methods and techniques. Some examples of these methods and techniques are reliability block diagram, network diagram, fault tree analysis (FTA), failure modes and effective analysis (FMEA), hazard rate and operability (HAZOP), etc. The implementation of these methods and techniques depends on factors such as the specific need, the tendency of the systems, etc. The most commonly used techniques in systems are reliability block diagrams. Reliability block diagrams which usually corresponds to the physical arrangement of items in the system. It is often used to model the impact of item failures on system performance. The major advantage of using the reliability block diagram approach is the ease of reliability expression and evaluation. In this section, only reliability block diagrams are explained. Using hazard rates of components, usage time of components and configuration (series, parallel) of system we may estimate reliability of robots. General information about the reliability block diagram representation is taken from references [32] and [33].

**Series System:** A reliability block diagram is in a series configuration when the failure of any block results in a system failure. Accordingly, for the functional success of a series system, all blocks must operate successfully during the intended task period of the system. Figure 1 shows N series reliability block diagram.

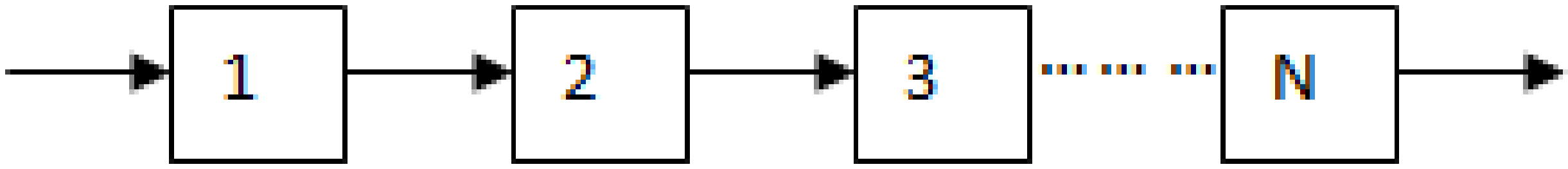

Figure 1: Series system reliability block diagram

System reliability $R_s(t)$;

$$R_s(t) = R_1(t) * R_2(t) \ldots * R_N(t) = \prod_{i=1}^{N} R_i(t) \quad (13)$$

where $R_i(t)$ represents the reliability of the ith block.

**Parallel System:** In a parallel configuration, only failure of all blocks results gives system failure. Accordingly, the success of only one block will be sufficient to guarantee the success of the system. Figure 2 shows N block parallel system block diagram.

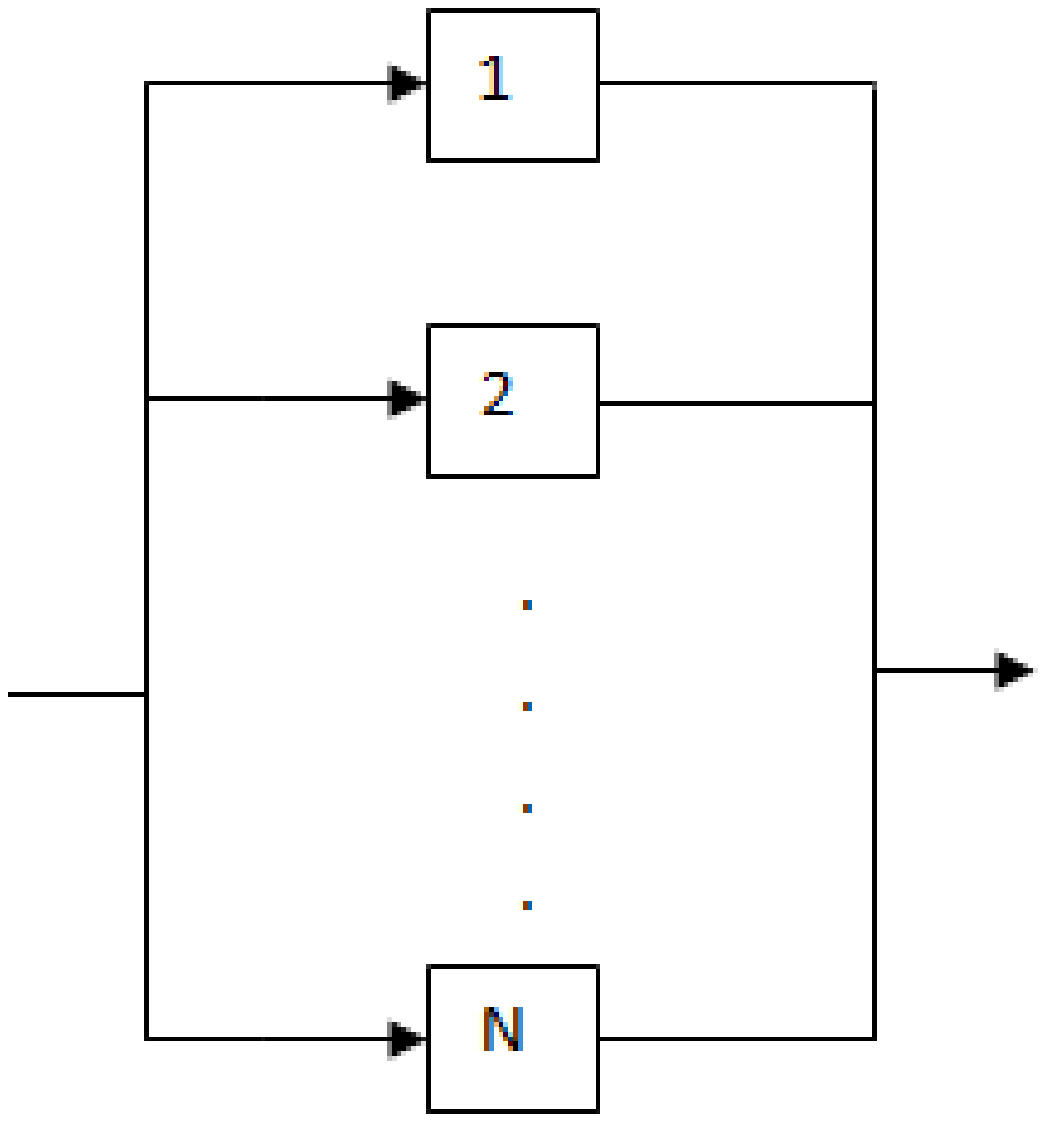

Figure 2: Parallel system block diagram

For a set of N independent blocks;

$$F_s(t) = F_1(t) * F_2(t) \ldots * F_N(t) = \prod_{i=1}^{N} F_i \quad (14)$$

Since eq. (15)

$$R_i(t) = 1 - \prod_{i=1}^{N} F_i \quad (15)$$

### 3.2.2 Life Distribution Models

Life distribution models required for the model-based method were obtained. Thus, by selecting the appropriate model, the reliability of the system with known failure rate will be obtained.

There are six different life distribution models in the literature such as Exponential, Weibull, Gamma, Normal,

Lognormal and Extreme value [32]. However, "Exponential" and "Weibull" life distribution models were used in the developed PHM tool within the scope of this study. Therefore, only Exponential and Weibull life distribution models are explained in this section. For other life distribution models, you can review the reference [32].

In this section, we discussed model-based methodology and most commonly used life distribution models are given to base to the reliability model. This section is used to calculate reliability of systems.

At initially, it is useful to look at some metric definitions that is given in Table 15 [34].

Table 15: Some metric definitions

| **Metric** | **Definition** |
|---|---|
| f(t), the probability density function | $P(a < t < b) = \int_a^b f(t)dt$ , to evaluate the likelihood of a failure occuring in the interval (a, b) |
| F(t), the cumulative distribution function | $P(t < b) = \int_0^b f(t)dt$, to evaluate the likelihood of a failure occuring by time b. |
| R(t), the survival function | $P(t > b) = 1 - F(b) = \int_b^\infty f(t)dt$ , to evaluate the likelihood of survival to time b. |
| $\lambda$ (t), the hazard function or instantaneous failure rate | $\lambda(t) = \frac{f(t)}{R(t)}$ |

**Exponential Distribution**: It is most commonly used in reliability analysis. This distribution can be attributed to primarily simplicity (constant hazard rate model), which corresponds to a realistic situation. It is widely used in electronics and probabilistic modeling. Equations related with this distribution given below.

$$f(t) = \lambda \exp(-\lambda t) \quad (16)$$
$$F(t) = 1 - \exp(-\lambda t) \quad (17)$$
$$h(t) = \lambda \quad (18)$$
$$MTTF = \frac{1}{\lambda} \quad (19)$$

**Weibull Distribution:** All three regions of the bathub curve can be represented by Weibull distribution and it is commonly used in many basic components such as capacitors, relays, bearings, etc. Equations related with this distribution given below.

$$f(t) = \frac{\beta(t)^{\beta-1}}{\alpha^\beta} exp\, exp\left[-\left(\frac{1}{\alpha}\right)^\beta\right] \quad (20)$$
$$F(t) = 1 - exp\, exp\left[-\left(\frac{1}{\alpha}\right)^\beta\right] \quad (21)$$
$$h(t) = \frac{\beta}{\alpha} exp\, exp\, (\frac{1}{\alpha})^{\beta-1} \quad (22)$$
$$MTTF = \alpha\, \Gamma\left(\frac{1+\beta}{\beta}\right) \quad (23)$$

Features of Weibull distribution over a range of $\beta$ values are given in Table 16 [32]:

Table 16: Features of Weibull distribution over a range of β values

| $\boldsymbol{\beta}$ | **Featrues** |
|---|---|
| $< 1.0$ | Decreasing failure-rate |
| 1.0 | Exponential (constant failure rate) |
| $> 1.0$ | Increasing failure-rate |
| 2.0 | Rayleigh single peak (linearly increasing) |
| ~3.5 | Normal shape |
| $> 10$ | Type I extreme value |

### 3.2.3 **Reliability Estimation**

Reliability generally shows how successful a system is in performing its intended function. In order to increase the system performance, intensive studies are carried out on reliability models and analytical tools. It is true that the consequences of unreliability in engineering can be very costly and often tragic. In a changing world, reliability becomes more important to ensure that systems operate with high performance and gain functionality. Reliability engineering is applied in many fields such as defense industry, robotics industry, aerospace industry etc. The current generation of mobile robots have poor reliability. In order to design more reliable robots, we need analytical tools for predicting robot failure. Reliability is the probability of the system performing its intended function under the specified operating conditions for a specified period of time. Mathematically, the reliability function R(t) (See eq. 13) is the probability that a system will work successfully without failure in time intervals between 0 and t [31].

$$R(t) = P(T > t), t \leq 0 \tag{24}$$

where T is random variable representing the failure the probability that a failure was caused by a component type c is simply in eq. (14).

$$P(failure) = \frac{Number\ of\ Failures\ Caused\ by\ c}{Total\ Number\ of\ Failures} \tag{25}$$

Reliability and failure are related as;

$$R(t) = 1 - F(t) = P(T \leq t) \tag{26}$$

where R(t) is the reliability function, and F(t) is the unreliability function. The failure rate function (hazard function), is very important, because it specifies the rate of the system aging and also shows rate of change over the life of a component. Failure rate function is given in eq. (27)

$$\lambda(\mathrm{t}) = \frac{R(t) - R(t + \Delta_t)}{\Delta_t R(t)} = \frac{f(t)}{R(t)} \tag{27}$$

where, $\lambda(t)dt$ indicates the probability that a device with age t will fail at small interval from time t + dt. If the hazard rate function follows an exponential distribution with parameter λ, the hazard rate function;

$$\lambda\ (\mathrm{t}) = \lambda \tag{28}$$

This means that the hazard rate function of the exponential distribution is constant which means the system does not have an aging property. The reliability function and the hazard rate are related as (if lifetime distribution function is exponential with parameter $\lambda$);

$$R(t) = e^{-\int_0^t h(x) \times dx} = e^{-\lambda t} \tag{29}$$

MTTF is the expected value of the reliability and related formulas about reliability, failure rate and MTTF are given in eq. (30) and eq. (31) respectively.

$$MTTF = \int_0^{\infty} R(t) \times dt \tag{30}$$

$$MTTF = \frac{1}{\lambda} \tag{31}$$

### 3.3 Probability of Task Completion Calculation (POTC)

Long-term autonomous operations of the robot is very important for the industrial applications. Reliability is very important concept to sustain autonomy in these industrial areas. Mobile robot's reliability depends on the subsystem, components and parts failure rate, usage time of these components, and environmental conditions. In order to increase the lifetime of mobile robots, prognostics-aware systems can be used. At these systems, robot should have known reliability and task completion probability. The probability of task completion (PoTC) is a metric that shows how successful mobile robots have completed a given task. PoTC could be used for lifetime extension of the system, sustaining autonomy, task allocation for multi-robot systems etc. by using reliability of robot and distance travelled along task. The lifetime of the robot maybe increased if the robot is capable of knowing health status and making task selection decision accordingly. Thus, using the reliability during decision making results sustainable autonomous operations for the robot. Probability of task completion (PoTC) for the robot is calculated by eq. (32).

$$PoTC = R^d \tag{32}$$

Where R and d are reliability of robot and total distance travelled by robot for given task respectively.

## 4. Prognostics and Health Management Tool

PHM Tool consists of 2 tabs. Different operations are performed on each tab in the PHM Tool. The operations performed on each tab are described in detail in this section.

Basically 2 different operations are performed in the PHM Tool:

- Robot Configuration
- System Analysis and Monitoring

Robot configuration operations are performed in "Robot Configuration Setup" tab. The flowchart of the operations performed on the "Robot Configuration Setup" tab is shown in Figure 3.

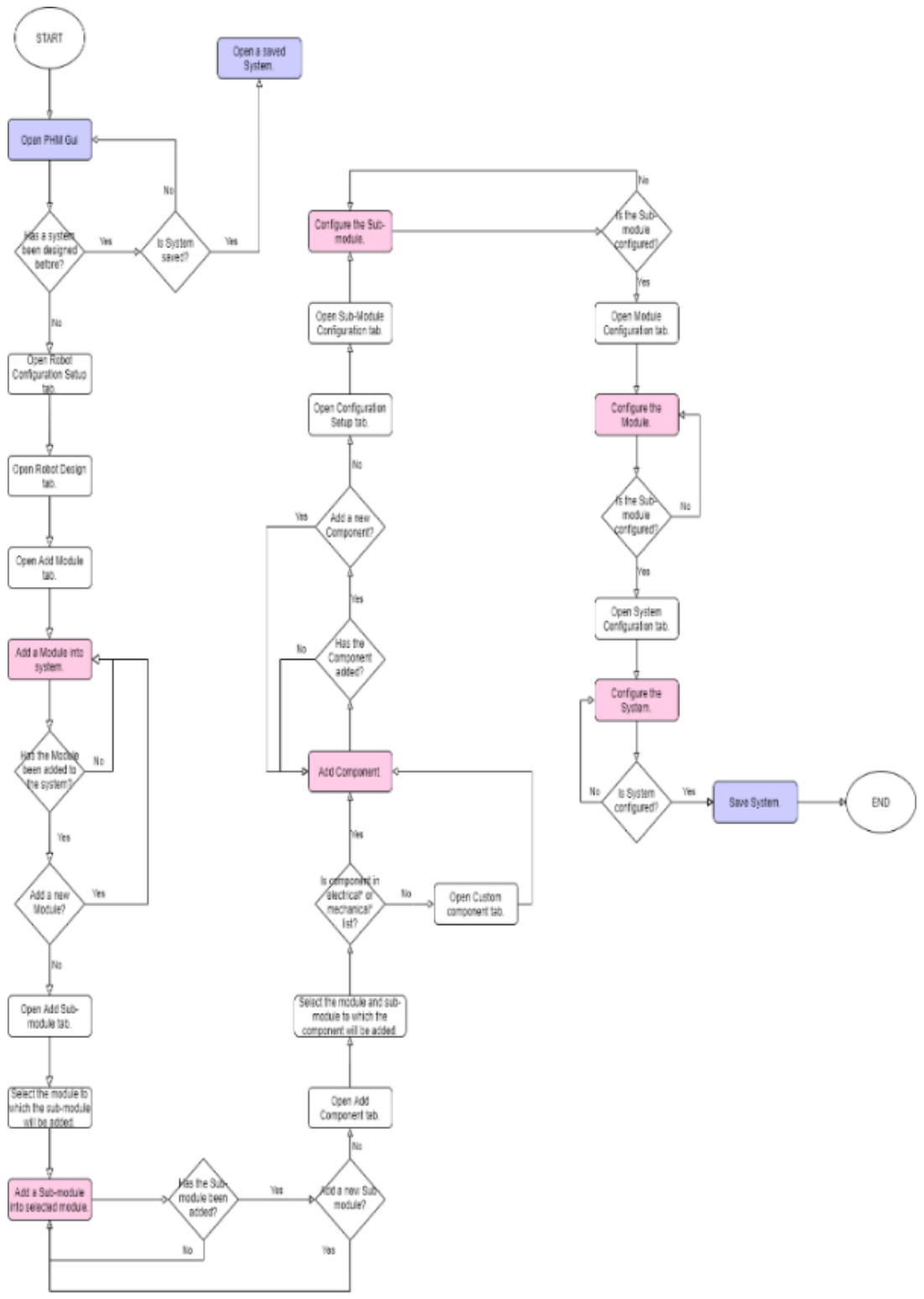

Figure 3: Flowchart of "Robot Configuration Setup" tab

System analysis and monitoring operations are performed in "Monitoring and Analysis" tab. The flowchart of the operations performed on the "Monitoring and Analysis" tab is shown in Figure 4.

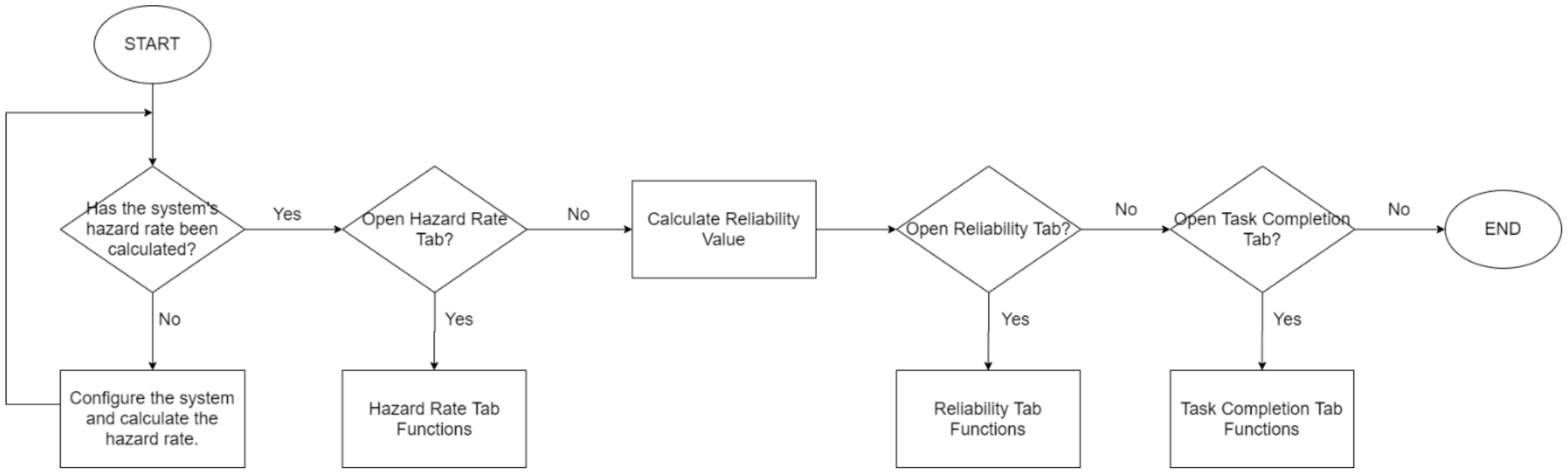

Figure 4: Flowchart of "Monitoring and Analysis" tab

PHM tool is model-based user interface which the user creates his system using various mechanical and electrical equipments, but can also calculates the reliability, failure rate and probability of task completion (POTC) of the created system. In addition, the PHM Tool offers the user the ability to formulate his own components and add them to the system. PHM Tool can also work with a real robot. Data from the sensors on the real robot are published via ROS topics. By subscribing to these topics in the PHM Tool, the system's failure rate, reliability and POTC values are calculated together with the data which is obtained from the sensors.

PHM Tool is a modular interface. The user can create his own system using various electrical and mechanical equipments with this tool. Failure rate values of equipments are calculated using hazard rate calculation algorithms [30-34-35-36] in PHM Tool. Electrical and mechanical equipments in the PHM Tool are given in Table 1. Failure rate calculation formulas of all equipments in this table are included in references [30-34-35-36]. Since the PHM Tool is a model-based tool robot design must be done in tool. Therefore modules, sub-modules and components in the system must be added by user into the tool. Design of the system is done in the "Robot Design" tab of the PHM Tool (Figure 5).

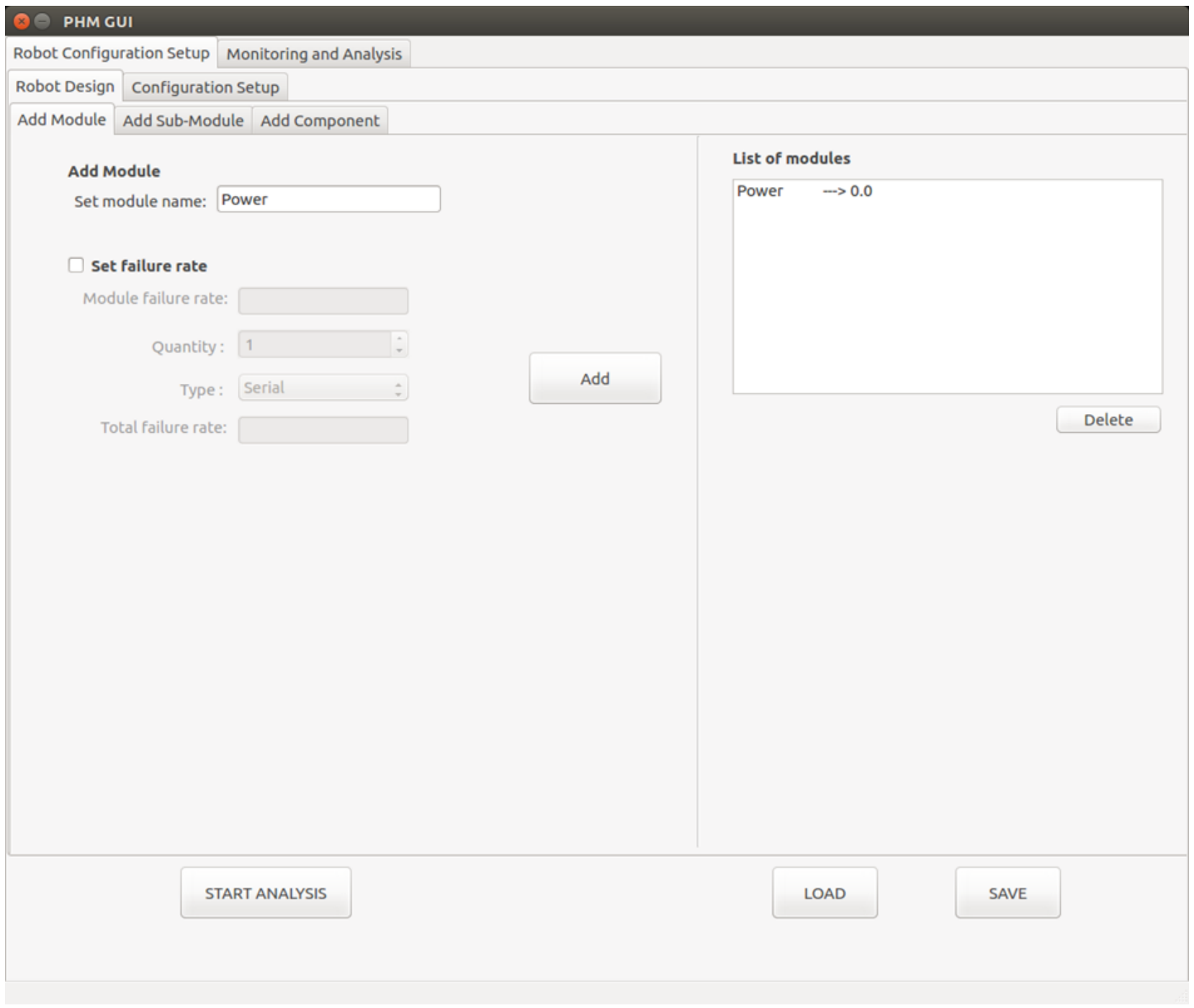

Figure 5: Robot Design Tab in PHM Tool

In PHM Tool, the reliability of the system has been calculated using the “Reliability Estimation Algorithms” which is located in Section 3 of the PHM Tool Model Based [37-38] document. The reliability of the system is calculated by the physical arrangements of each item that creates the system. Thus, the effect of the failure rate value of each item on the system is modeled.

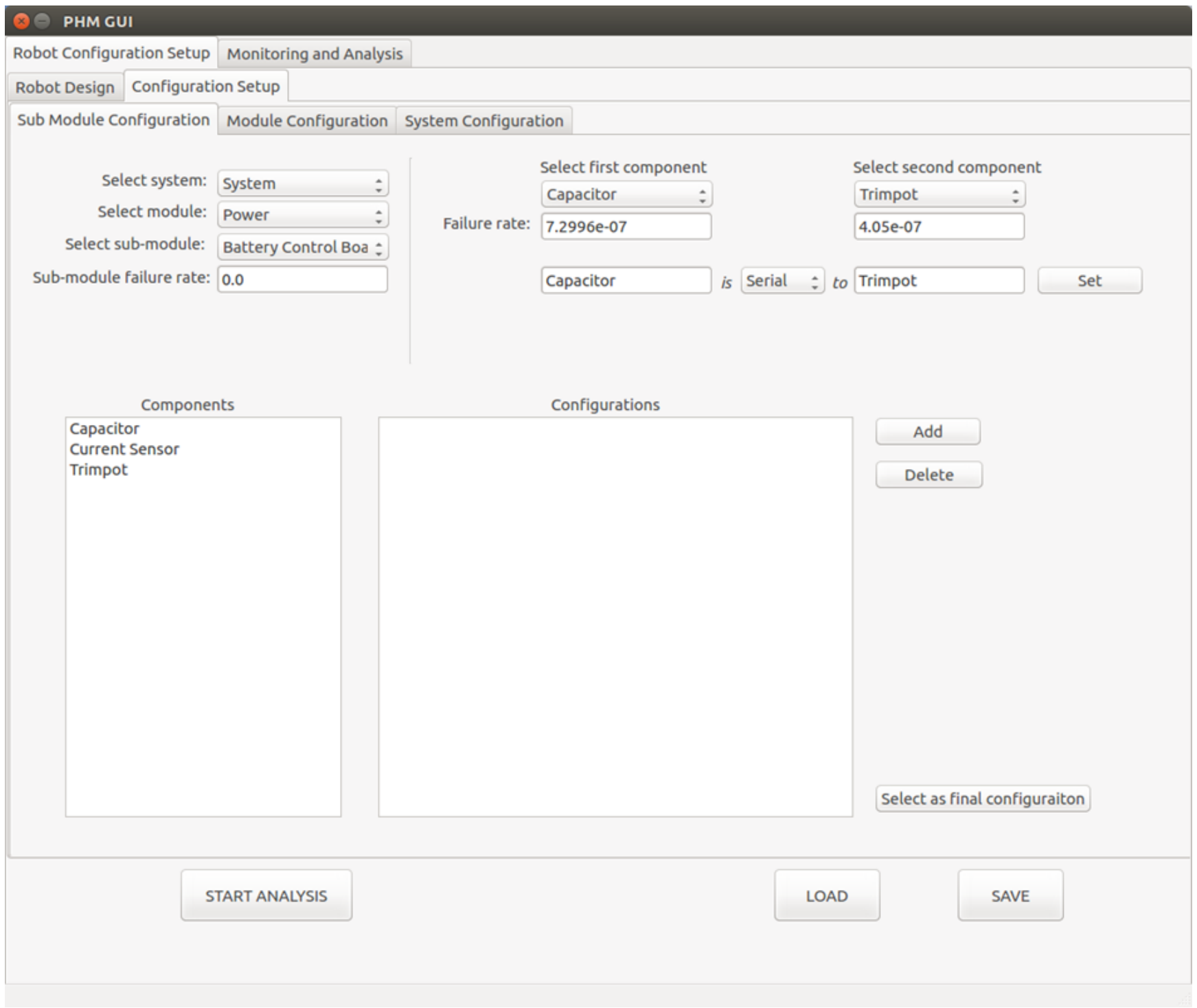

Figure 6: Configuration Setup tab in PHM Tool

Physical arrangements of the items in the system are made in 2 ways in the PHM Tool:

- Series
- Parallel

In a series configuration when the failure of any item results in a system failure. Accordingly, for the functional success of a series system, all items must operate successfully during the intended task period of the system. In a parallel configuration, only failure of all items results give system failure. Accordingly, the success of only one item will be sufficient to guarantee the success of the system. Configuration of the system is done in “Configuration Setup” tab in PHM Tool (Figure 6). How to configure the system is explained in previous study of PHM Tool User Guide 3.3.2 section with given document [39].

Accelerated life test (ALT) models given which is used in different conditions in order to calculate reliability analysis in this tool. Formulas for ALT models used in the interface are located in Section 4 of the PHM Tool Model Based [37] document. The user can perform reliability analysis by choosing Exponential and Curve Distribution models. Equation (29) is used for reliability calculation and equation (32) is used for probability of task completion (POTC). In equation (32), the unit of time (t) in the formula used to calculate the reliability of the system should be hour or second. The unit of distance (d) in the formula used to calculate the POTC should be km or m. In PHM Tool the user has the opportunity to select any of these units (Figure 7). When the user chooses any of these units (m/s or km/h), reliability and POTC values are calculated according to the selected units.

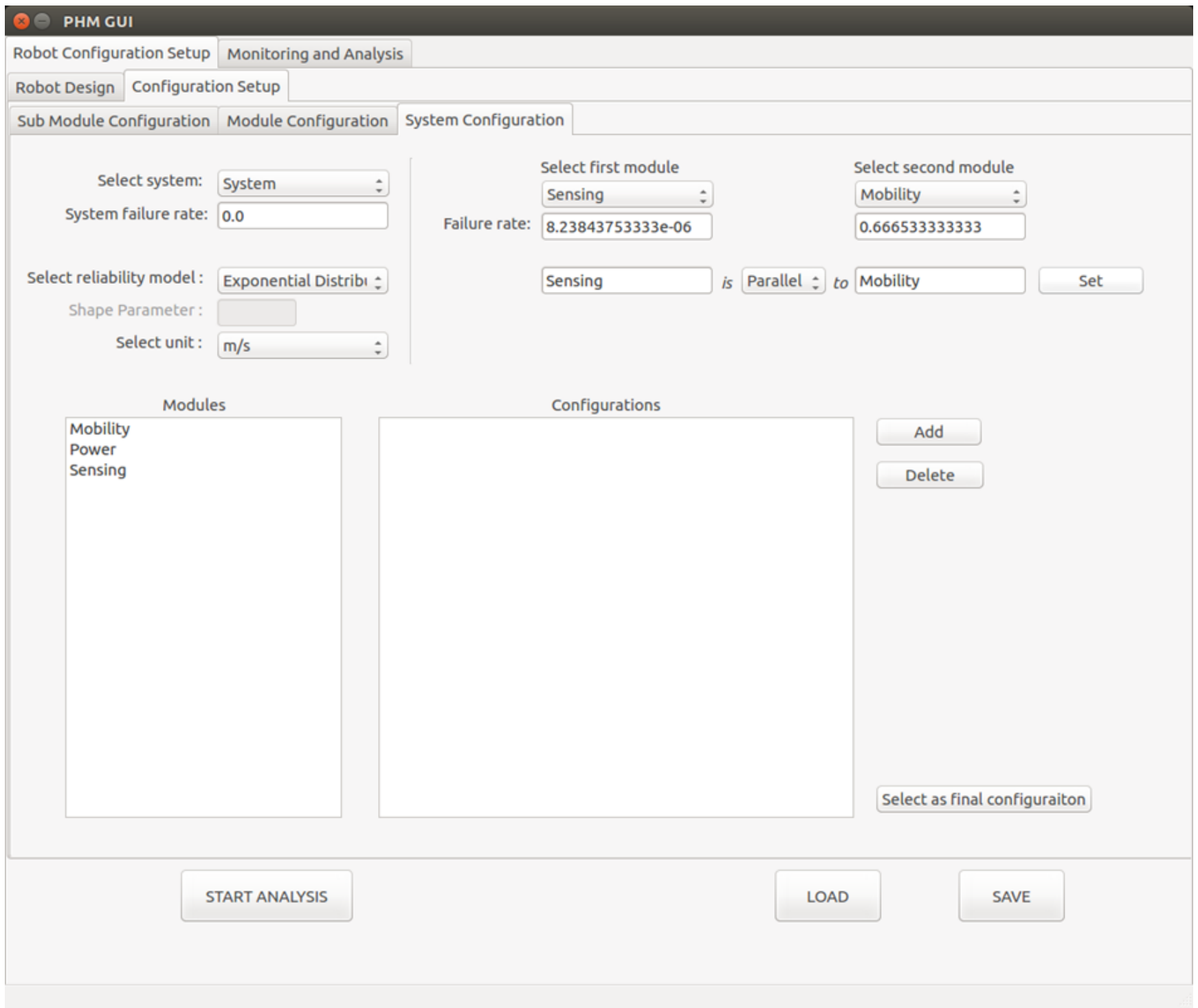

Figure 7: ALT Models and unit selection in System Configuration tab

After the system is configured, the system must be started for analysis by clicking the “Start Analysis” button. There are 3 types of analysis available in PHM Tool:

- Hazard Rate Analysis
- Reliability Analysis
- Task Completion Analysis

How to analyze the system is explained in Section 3.4 of PHM Tool User Guide [39] document.

In the “Hazard Rate Analysis” and “Reliability Analysis” tabs (Figures 8 and 9), the graphs of the calculated hazard rate and reliability values are plotted in real time. At the same time, the user can add the sensor data that are read from ROS topics in to the appropriate modules. Thus, besides the nominal values, a sensor-based hazard rate and reliability values should be calculated.

**Nominal:** The calculated hazard rate and reliability value of the system without adding any sensor.

**Sensor based:** Hazard rate and reliability values calculated by adding sensors to the modules in the system.

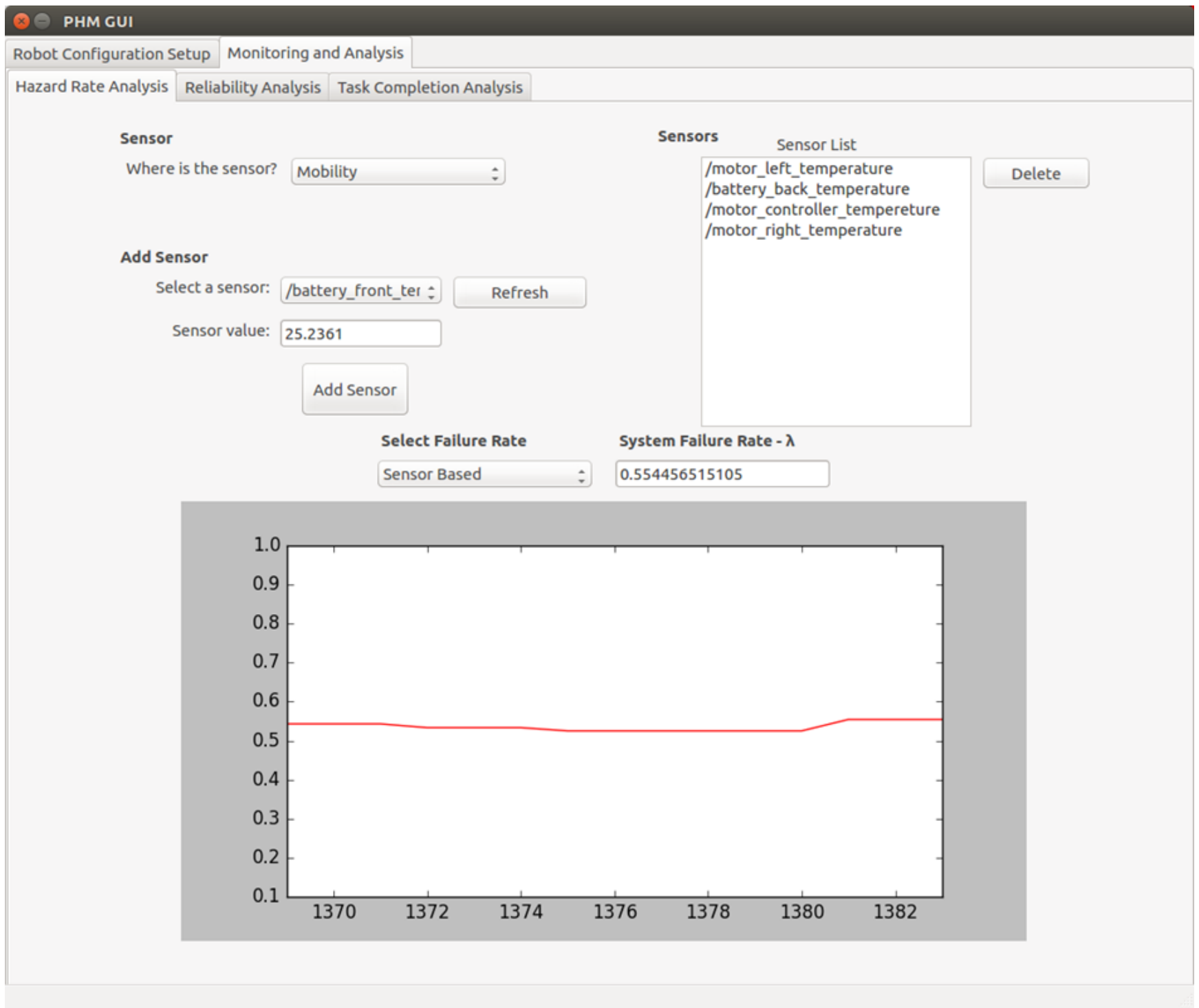

Figure 8: Hazard Rate Analysis tab in PHM Tool

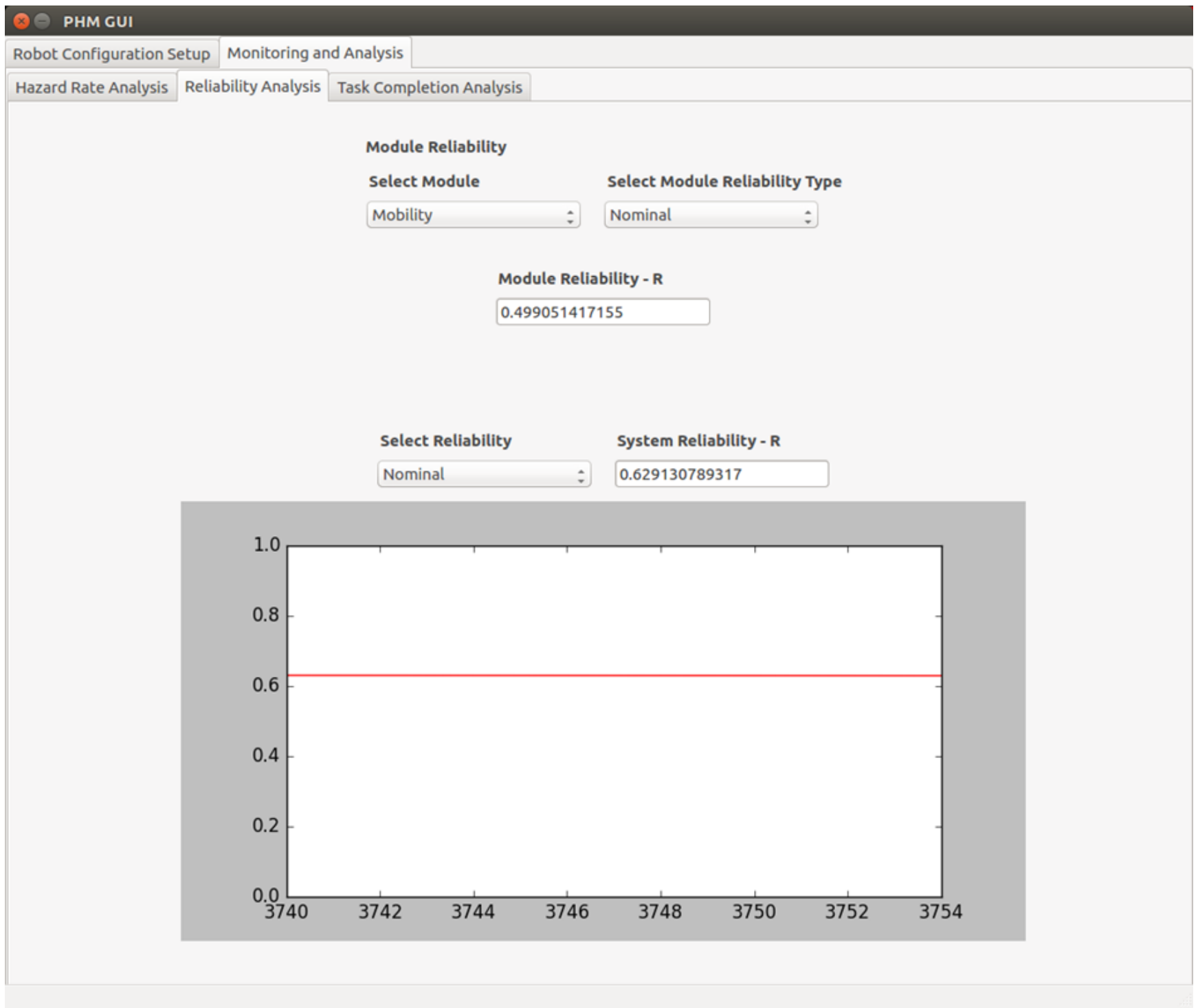

Figure 9: Reliability Analysis tab in PHM Tool

In the "Task Completion Analysis" tab (Figure 10), the calculated reliability value is used. In order to calculate the POTC value, the task sent for the mobile robot must come to PHM Tool. POTC value, total time and distance values are estimated before realizing the robot task by applying Predict analysis as soon as the task arrives. After estimation, besides the total time and distance values, nominal and sensor based POTC values are calculated and displayed on the interface and POTC values are added graphically. In addition, the incoming task can be used to calculate advanced simulation by adding it to the list of tasks in the "Prognostic Analysis" tab (Figure 11). The robot transmits distance and time values to PHM Tool after finishing the current task. POTC value is calculated as nominal and sensor based by applying actual analysis. The calculated POTC values are displayed in the interface along with the total time and distance values, and POTC values are added graphically.

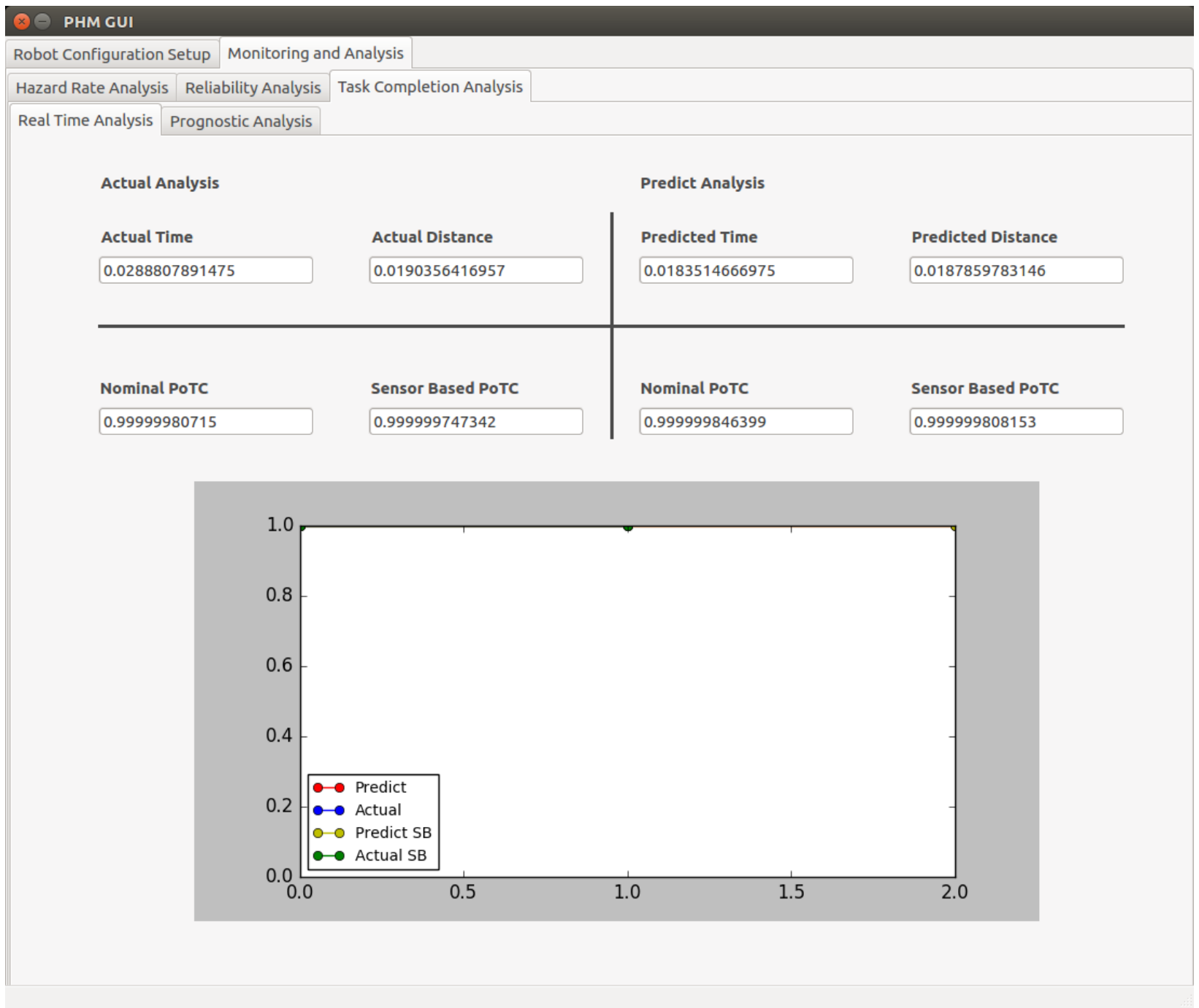

Figure 10: Task Completion Analysis tab in PHM Tool

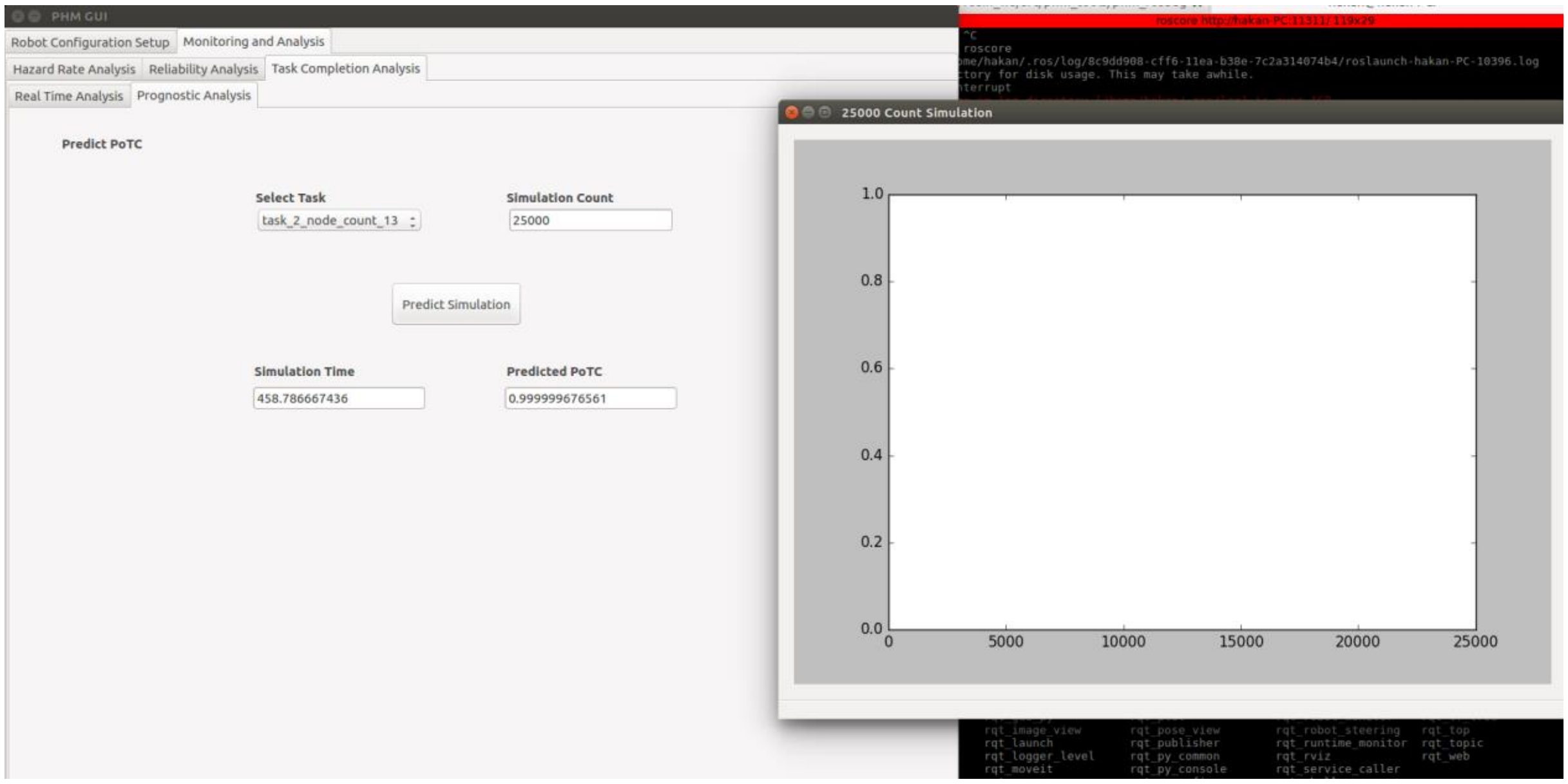

Figure 11: POTC Prognostic Analysis Tab in PHM Tool

# 5. Case Study for PHM Tool

In following sections, in 5.1 OTA design and configuration is given in short. In 5.1.1. OTA design and in 5.1.2 OTA configuration is explained. After that, in section 5.2 case study implementation is given in details.

### 5.1 OTA Design and Configuration

Since PHM Tool is a modular tool, each of the modules, submodules and components in OTA must be added in this tool. OTA design is done by adding modules, submodules and components in PHM Tool.

OTA should be configured by making physical arrangements of these added parts according to each other. As a result of the design and configuration of OTA in the PHM Tool, the hazard rate and failure rate values of the system are calculated.

#### 5.1.1 OTA Design

The modules, sub-modules, components on the OTA, their quantities and failure rate values calculated for one of them with the PHM Tool are shown in Table 17. OTA design is done in the "Robot Design" tab under the "Robot Configuration Setup" tab in the PHM Tool.

The procedures for adding all the equipment in OTA to the PHM Tool are described in Section 3.1 in the PHM Tool Use Case [40] document. At the same time, the design processes for the OTA use case scenario of PHM Tool are explained in the video on the PHM Tool - Robot Design link [43].

Table 17: Modules, sub-modules and components list of OTA with component failure rates

| Module | Sub-module | Component | Quantities | Failure Rates |
|---|---|---|---|---|
| **Power** | Battery | - | 4 | 2,00E-08 |
| | Battery Control Board | Capacitor | 28 | 9,36E-06 |
| | | Thermistor | 9 | 7,33E-06 |
| | | Diode | 16 | 6,25E-07 |
| | | LED | 9 | 5,00E-06 |
| | | Resistor | 26 | 1,33E-07 |
| | | Trimpot | 3 | 2,42E-06 |
| | | Step Down Switching Regulator | 7 | 3,96E-07 |
| | | Terminal Blocks/Connector | 9 | 2,21E-06 |
| | | Connector/Socket | 1 | 1,18E-07 |
| | | Inductor | 7 | 5,80E-08 |
| | | Current Sensor | 2 | 2,50E-07 |
| | | Female Header | 1 | 1,90E-06 |
| | Low Level Control Unit: DSK-MD | XT60 Socket | 4 | 7,53E-08 |
| | | Zener Diode | 4 | 1,77E-07 |
| | | Ultrafast Diode | 4 | 5,29E-07 |
| | | Aluminum Capacitor | 4 | 1,40E-07 |
| | | Micro USB Connector | 1 | 6,07E-07 |
| | | 16MHz Cryst. Osc. | 1 | 1,77E-06 |
| | | 32.768Khz Cryst. Osc. | 1 | 4,26E-07 |
| | | Resistors | 65 | 4,99E-07 |
| | | STM32F407VGT6 | 1 | 1,00E-08 |
| | | Resettable Fuse | 2 | 2,30E-06 |
| **Sensing** | SICK Laser Sensor | - | 1 | 8,00E-08 |
| | IPS | Antenna | 1 | 5,80E-07 |
| | | Micro USB Connector | 1 | 6,07E-07 |
| | | DW1000 Chip | 1 | 8,60E-08 |
| | | Resistor | 5 | 4,46E-07 |
| | | Capacitor | 5 | 9,36E-07 |
| | Camera | - | 1 | 5,88E-09 |
| **Communication** | Communicaton Card: DSK-M | Resistor | 9 | 4,53E-07 |
| | | Capacitor | 9 | 1,66E-07 |
| | | 25MHz Crystal Oscillator | 1 | 6,54E-06 |
| | | 16Mhz Crystal Oscillator | 1 | 5,90E-07 |
| | | STM32F407VGT6 | 1 | 2,30E-07 |
| | | SN65HVD230DR | 1 | 1,70E-07 |
| | | Fuse | 4 | 8,00E-08 |
| | | LAN8720A-CP | 1 | 5,90E-07 |
| | | Micro USB Connector | 1 | 2,43E-06 |
| | Antenna | - | 1 | 7,50E-07 |
| **Mobility** | Encoder | - | 2 | 5,80E-06 |
| | Driver wheel | - | 2 | 1,80E-07 |
| | Caster wheel | - | 4 | 6,80E-06 |
| | DC Motor | - | 2 | 8,04E-06 |
| | Bearing | - | 4 | 3,61E-11 |
| **Computation** | YSK-M: High Level Control Unit | Inno-Box Industrial Box PC | 1 | 6,50E-08 |
| | YSK-G: Vision Control Unit | NVIDIA JETSON TX2 | 1 | 5,00E-08 |

### 5.1.2 OTA Configuration

After each equipment in the OTA is added in the "Robot Design" tab of the PHM Tool, the system must be configured in order to calculate the total failure rate and reliability values of the system. The configuration of the OTA is done in the "Configuration Setup" tab under the "Robot Configuration Setup" tab in the PHM Tool. Configuration of the equipment in the OTA is included in section 3.2 of the PHM Tool Use Case [40] document. At the same time, the configuration operations performed in the PHM Tool use case scenario for OTA are explained in the video in the PHM Tool - Robot Configuration link [44].

While configuring OTA, the following steps are followed in order:

- Configuring components under each submodule:

This section is explained in detail in Section 3.2.1 of the PHM Tool Use Case [40] document.

- Configuring submodules under each module:

This section is explained in detail in Section 3.2.2 of the PHM Tool Use Case [40] document.

- Configuring the modules under the system (OTA):

The configuration of the OTA is made as in the Figure 12. This section is explained in detail in section 3.2.3 of the PHM Tool Use Case [40] document.

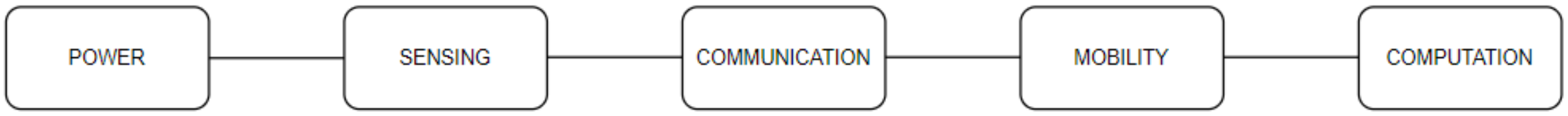

Figure 12: System (OTA) Configuration

## 5.2 Case Study Implementation

In the Monitoring and Analysis tab, the configured system in the “Robot Configuration Setup” tab is presented to the user in real time by calculating the hazard rate, reliability and POTC values according to the specified reliability model and unit. In addition, sensors can be added and the results are shown in real time. Monitoring and Analysis is explained in detail in Section 4 of the PHM Tool Use Case document [40].

The Gazebo environment [45] used in the use case is shown in Figure 13. Tasks coming to OTA are monitored on this Gazebo environment. The movement operations it performs on OTA, Gazebo and the objects detected by the sensor are visualized on Rviz in Figure 14.

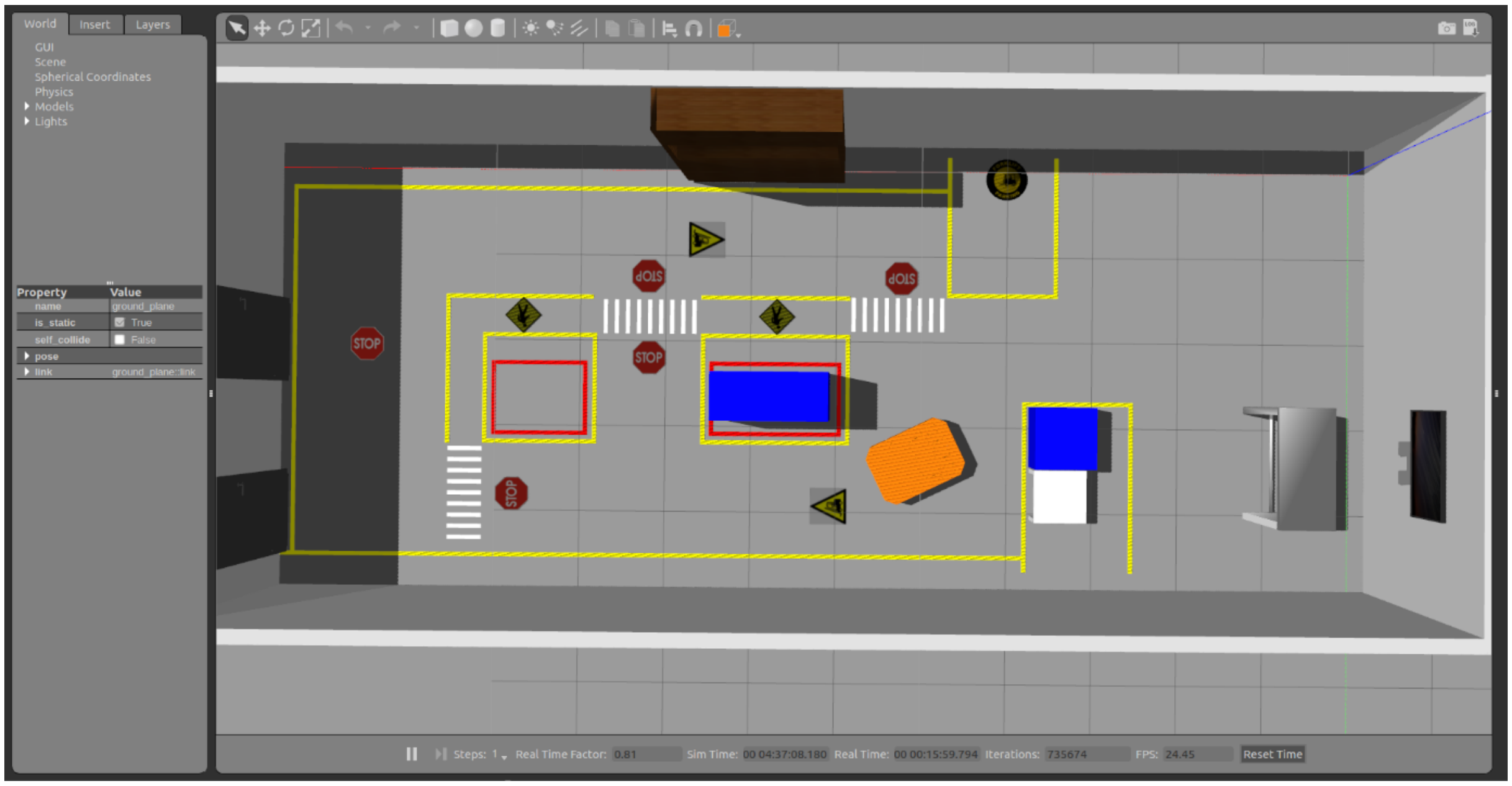

Figure 13: Gazebo Environment

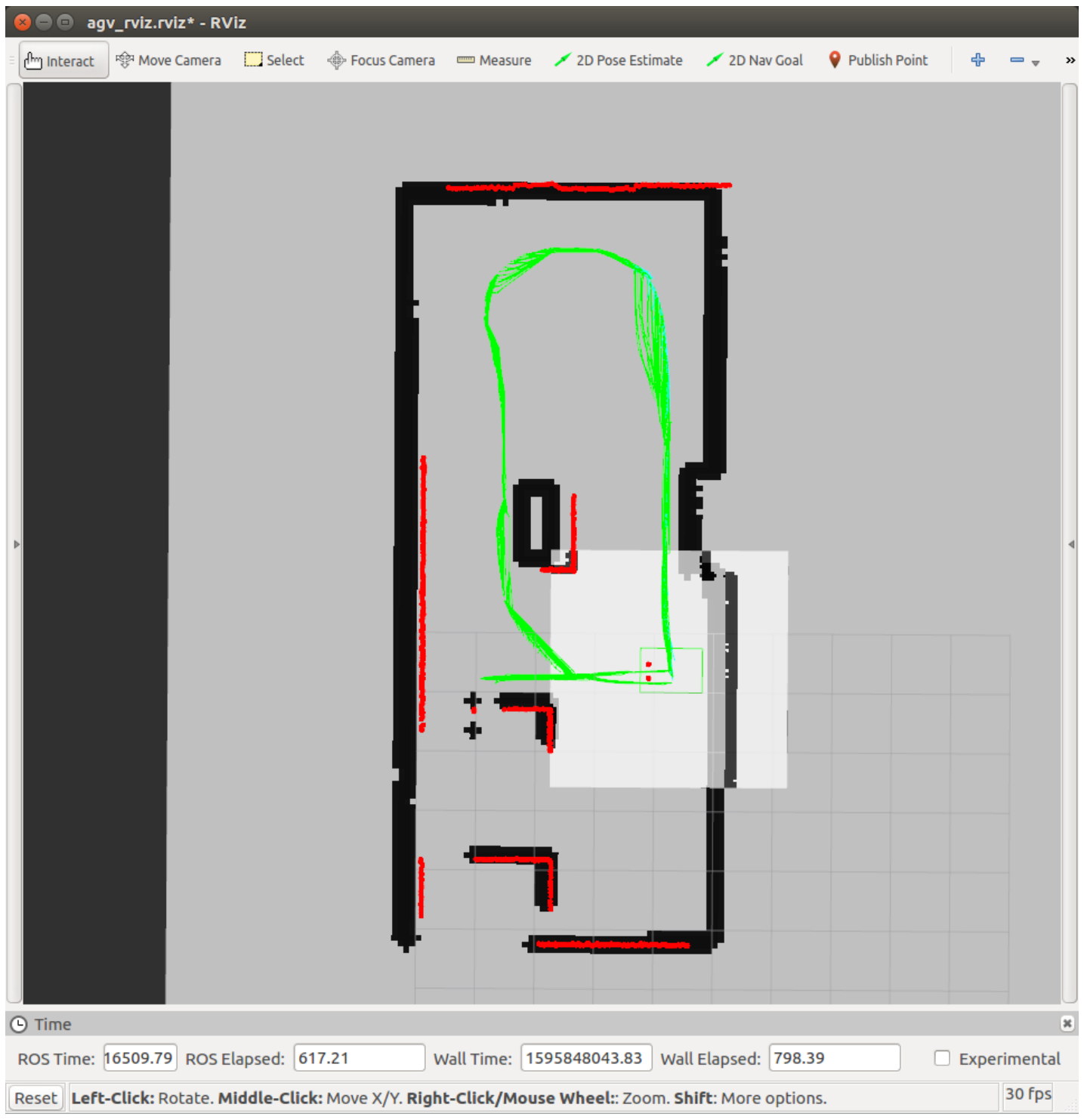

Figure 14: Rviz

SMACH [46] is a Hierarchical State machine builder tool that gives advantage on code introspection and fast prototyping. In case study, SMACH takes the tasks received from the server and applies the appropriate states in the state machine architecture, enabling OTA to successfully perform the incoming tasks. States created in SMACH with SMACH Viewer are visualized as in Figure 15.

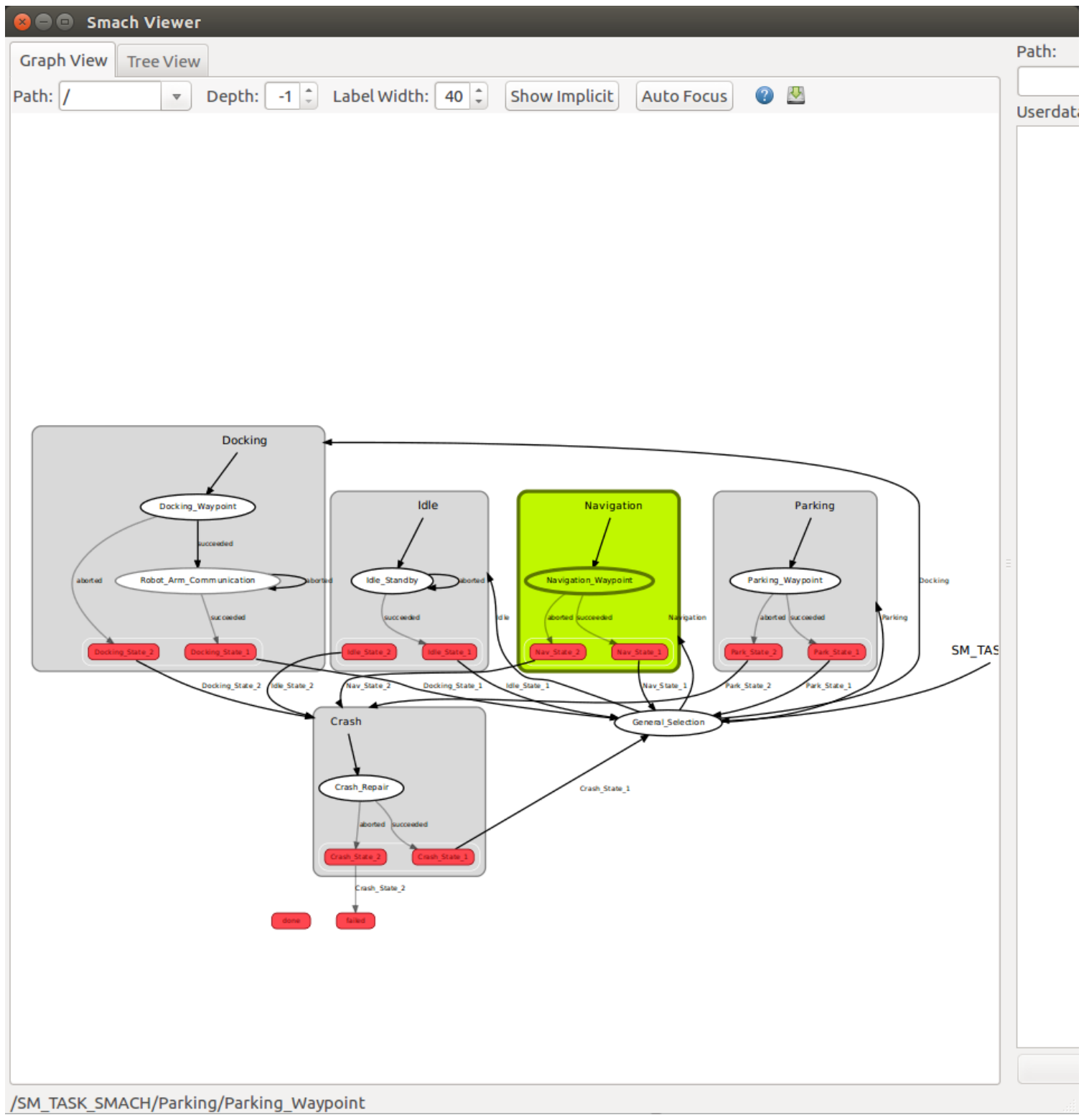

Figure 15: SMACH Viewer

Task information named task_time, task_positions and robot_task_list applied by SMACH are sent to PHM Tool for POTC calculation. This architecture created is shown in Figure 16.

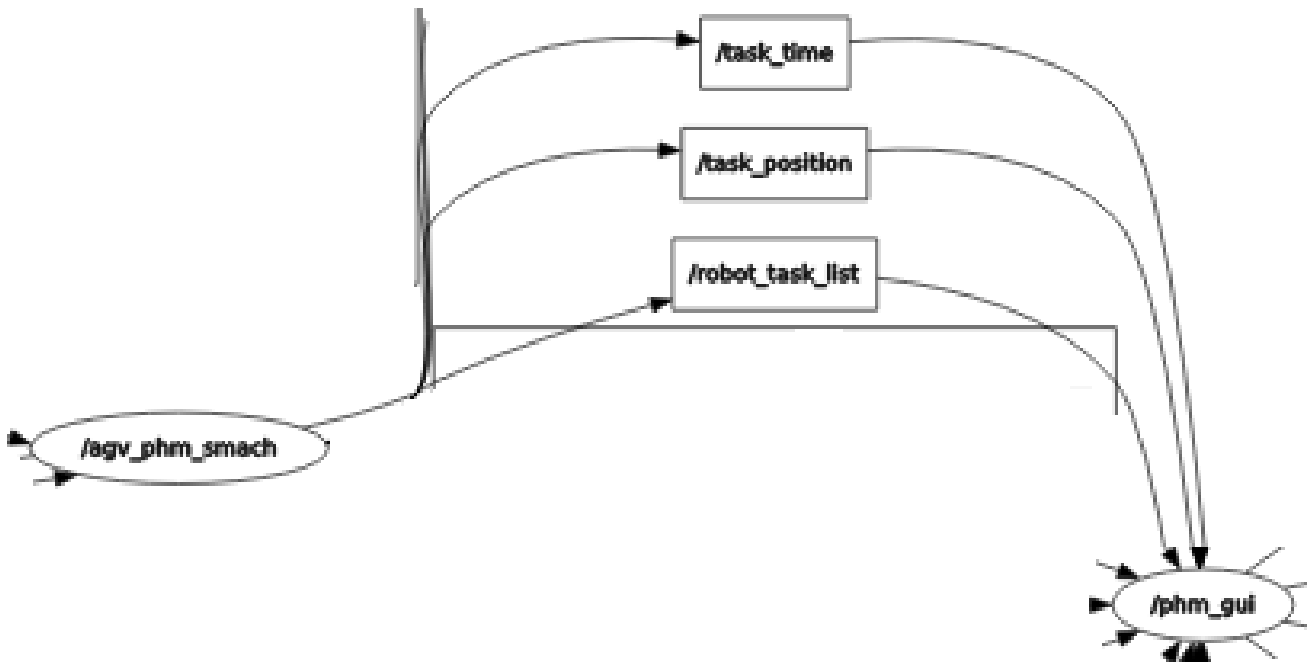

Figure 16: Publishing Tasks from SMACH to PHM Tools

In the case study, the sensor data read from the OTA were collected and recorded with rosbag. It calculates the Hazard Rate values by reading the sensor data selected from the PHM Tools interface over the rosbag. The architecture of this process is shown in Figure 17.

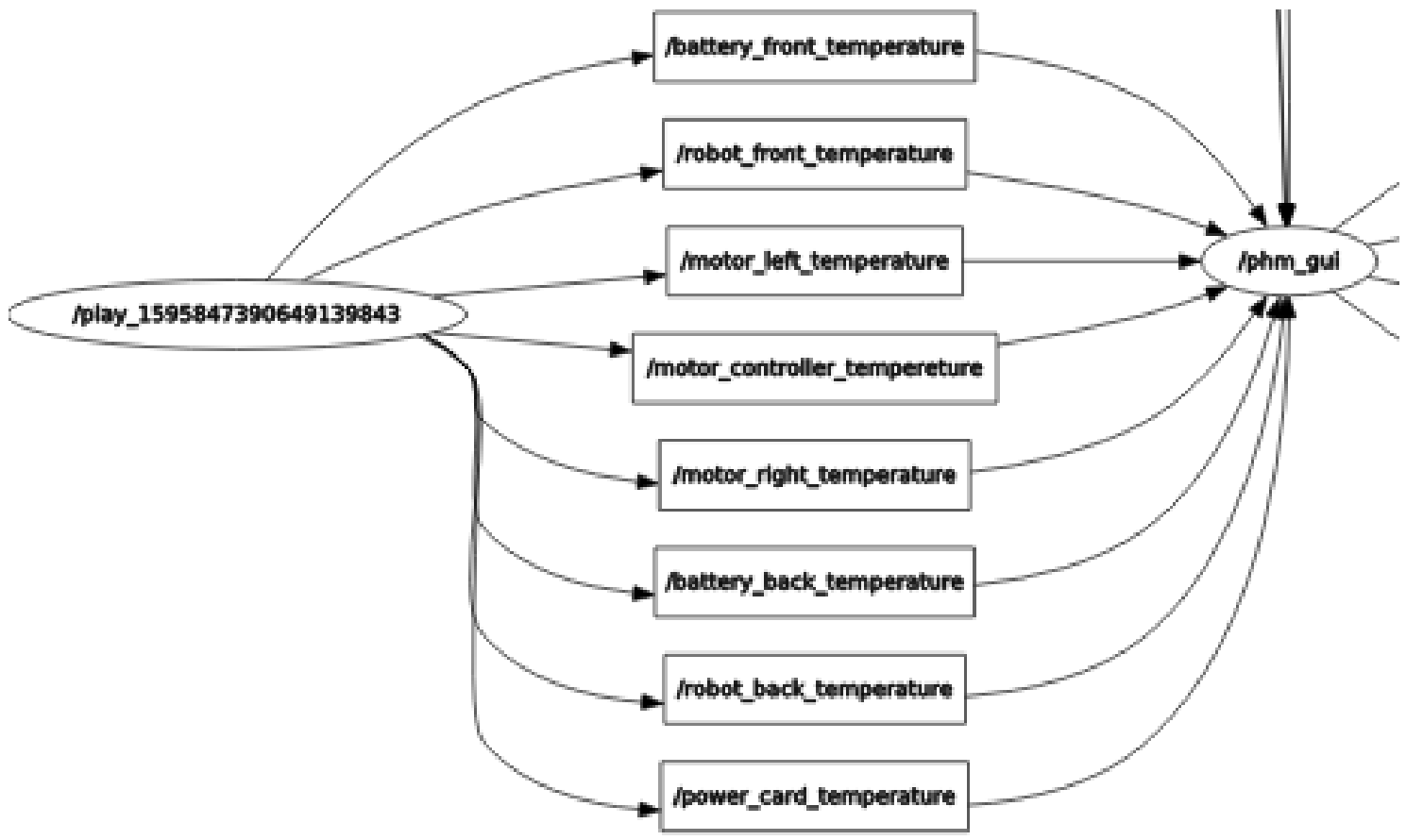

Figure 17: Sensors Added from Rosbag

PHM Tools applies filtering by sending all the calculated hazard rate, reliability and PoTC data to phm_hazard_rate_calculation_node, phm_reliability_estimation_node and phm_robot_task_completion_node nodes for users to use. The filtered data is then published in the phm_hazard_rate, phm_reliability and phm_potc topics for users to use. This architecture created is shown in Figure 18.

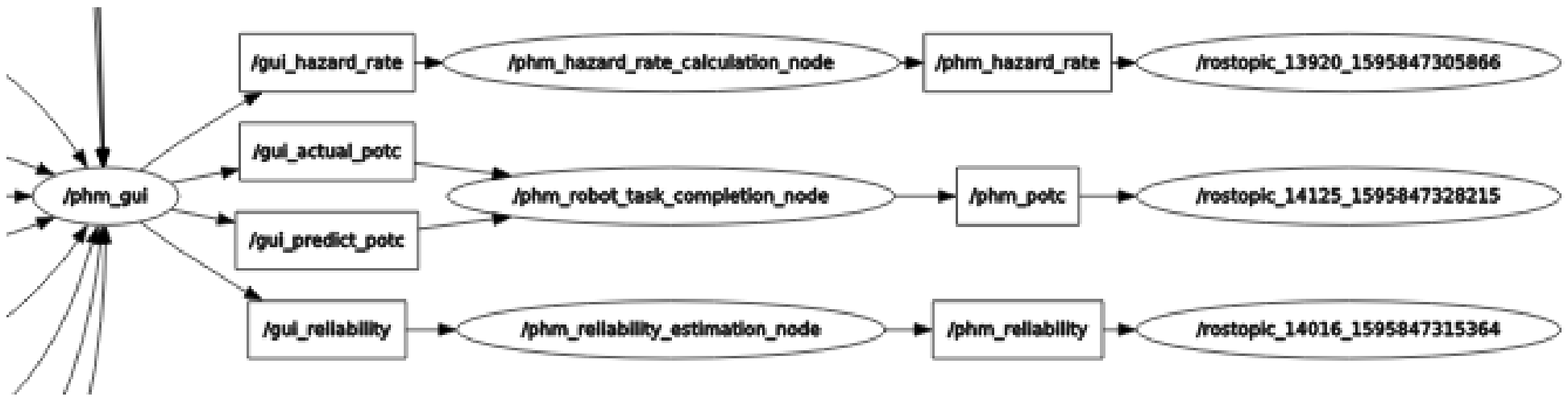

Figure 3 Architecture Between PHM Tools Nodes

## 6. Conclusion

In this paper, Prognostic and Health Management tool infrastructure which uses a model-based methodology for robotic systems, is given. The aim is developing general Prognostic and Health Management tool for ROS. In this aim, Importance of PHM in industry and robotic applications are given with comparison of the existing PHM tools and methods. Model for Health estimation and mathematical background explained initially and Prognostic tool is explained as a guide. In an Example scenario, OTA design and configuration is given in detail. reliability, remaining useful life (RUL), and the probability of task completion of the robot (PoTC) are estimated in this tool. Reliability and PoTC of the system are calculated by using hazard rate of the mechanical and electrical components, configuration of the robot (series, parallel, etc.). Results show that health of the system depends on hazard rate of all components in the system and their configurations, Also, usage time and some environmental conditions (temperature, load, humidity, etc.) affect system's overall condition. Proposed PHM tool can be obtained by given link (http://wiki.ros.org/phm_tools) and used in further robotic applications with following steps given in previous sections. With extending use cases, PHM tool will provide further applications in robotics.

**Acknowledgements**: This study is supported by ROS-Industrial Focused Technical Project (ROSin FTP) program (Project Name: Prognostics and Health Management Tool for ROS). Website: https://www.rosin-project.eu/ftp/prognostics-and-health-management-tool-for-ros